\documentclass[journal]{IEEEtran}
\usepackage{latexsym,,amssymb,amsmath,graphicx,epsf,cite,bbm,float}
\usepackage{ifpdf}
\usepackage{epstopdf,mathtools}

\usepackage{algorithm,algorithmic}
\usepackage{amsmath,amssymb,bm}
\usepackage{amsfonts,dsfont,color,bbm,subcaption}

\newcommand{\abs}[1]{|#1|}

\usepackage{hyperref,amsthm}

\newtheorem{theorem}{Theorem}

\newtheorem{lemma}[]{Lemma}

\newtheorem{proposition}[]{Proposition}

\newtheorem{remark}[]{Remark}

\begin{document}

\title{An Auto-Regressive Formulation for Smoothing and Moving Mean with Exponentially Tapered Windows} 
\author{\IEEEauthorblockN{Kaan Gokcesu}, \IEEEauthorblockN{Hakan Gokcesu} }
\maketitle

\begin{abstract}
	We investigate an auto-regressive formulation for the problem of smoothing time-series by manipulating the inherent objective function of the traditional moving mean smoothers. Not only the auto-regressive smoothers enforce a higher degree of smoothing, they are just as efficient as the traditional moving means and can be optimized accordingly with respect to the input dataset. Interestingly, the auto-regressive models result in moving means with exponentially tapered windows.
\end{abstract}

\section{Introduction}
Data smoothing is heavily utilized in many learning problems to detect certain patters in the data or mitigate noisy/anomalous values \cite{hardle1991smoothing}. The notion of smoothing stems from the inference that the data points close to each other in their features should have close mappings (functional evaluations), i.e., behave similarly. The most prominent example is a time-series dataset where data points closer in time should have similar values, e.g, unnaturally high or low values should be suppressed to match the nominal data pattern.
Smoothing is mainly done on the datasets for easier extraction of their information and consequently ease their analyses \cite{simonoff2012smoothing}. This approach is applied in many fields including signal processing \cite{roberts1987digital,orfanidis1995introduction,fong2002monte,schaub2018flow}, data analysis \cite{kenny1998data,brandt1998data,guo2004functional,deveaux1999applied,zin2020smoothing}, machine learning \cite{gokcesu2021regret,saul2004overview,gokcesu2020recursive,yaroshinsky2001smooth,gokcesu2020generalized,servedio2003smooth,gokcesu2021optimal,hartikainen2010kalman, gokcesu2021optimally}, and anomaly detection \cite{chandola2009anomaly,gokcesu2017online,tsopelakos2019sequential, gokcesu2018sequential}.
While curve-fitting is closely related to smoothing, it has a major distinction \cite{arlinghaus1994practical}. The curve-fitting aims to construct a well-defined function using the observed data points, e.g., interpolation, which is not of interest in smoothing. 

The most traditional way of smoothing is via linear smoothers, in which each data point is smoothed with linear weighting of nearby data points \cite{buja1989linear}. In other words, given an input $\boldsymbol{y}$ and linear weights or Finite Impulse Response (FIR) filter $\boldsymbol{w}$, the smoothed $\boldsymbol{x}$ is given by their convolution \cite{oppenheim1997signals}.

The most commonly utilized linear smoother (especially for smoothing of a time-series) is the moving (rolling) mean (average) filter, where the linear weights $\boldsymbol{w}$ are all non-negative and sum to $1$ (i.e., a probability distribution) \cite{chatfield2013analysis,wei2006time,gokcesu2018adaptive,hansun2013new}. 
Hence, the weights constitute a low-pass filter to mitigate the higher frequency deviations and smooth out the original input signal \cite{kaiser1977data}.
The moving mean smoother is applied to better detect long-term behavior or trend, where the emphasize is directly related to the filter length \cite{hatchett2010optimal}.
The use of the moving mean linear smoother is commonly encountered in mathematical finance to analyze features like the trading price and volumes in the stock market \cite{chiarella2006dynamic,gokcesu2021nonparametric}. Similarly, it is also used to analyze the trends of many macroeconomic features including import/export volume, unemployment statistics and gross domestic product \cite{schwert1987effects}. It is also utilized in biomedical signal processing such as electrocardiogram \cite{bai2004combination,chen2003moving}.

Furthermore, the smoothing of an input signal is not only achieved by the moving mean filter but also other diverse filters as well, such as the rolling median filters \cite{justusson1981median}. Note that the moving mean filter becomes statistically optimal in the maximum likelihood sense when the input signal is contaminated by Gaussian noise \cite{kalman1960new}. Nonetheless, when the contamination is non-Gaussian and has heavy tails in its distributions, the performance of the moving mean degrades substantially and becomes easily influenced by undesired outliers. More robust approaches are studied to this effect \cite{gokcesu2021generalized,khargonekar1985non,gokcesu2022nonconvex}, where a prominent example is the moving median filter. Although the moving median is a nonlinear smoother and the smoothed signal is acquired by the weighted median of close groups of samples, it is efficiently computable by use of indexable skiplist \cite{pugh1990skip}. We point out that specifically for the moving median, the smoothed signal is statistically optimal in the maximum likelihood sense, when the signal contamination results from a Laplace distribution \cite{arce2005nonlinear}. The moving median filter has many applications including mass spectrometry \cite{do1995applying}, edge detection \cite{huang1998computing}. The application is especially rich in the field of image processing because of the edge-preserving capabilities \cite{ataman1981some}. 

To recap, given the observed input $\boldsymbol{y}$, the smoothed output $\boldsymbol{x}$ is generally created by use of a moving operation $\mathcal{T}(\cdot)$, i.e.,
\begin{align}
	\boldsymbol{x}=\mathcal{T}(\boldsymbol{y}),
\end{align}
where $\mathcal{T}(\cdot)$ can be either the rolling mean, the rolling median or some other function. By incorporation of weighted window function $\boldsymbol{w}$, we can generalize the operation as
\begin{align}
	\boldsymbol{x}=\mathcal{T}(\boldsymbol{y},\boldsymbol{w}),
\end{align}
where the relevant operations instead become weighted moving mean or weighted moving median, etc. In this work, we specifically focus on the moving mean smoothing operation.

We point out that in this smoothing operation, the design boils down to the weighting, which are provided by specific window functions \cite{weisstein2002crc} that are heavily utilized in many fields including antenna design \cite{rudge1982handbook}, spectral analysis \cite{stoica2005spectral} and beam-forming \cite{van1988beamforming}.
We focus on their applications in statistical analysis and smoothing \cite{dixon1951introduction}. Note that, these weighting windows are also referred as kernel functions \cite{keerthi2003asymptotic} in Bayesian analysis \cite{ghosh2006introduction} and curve fitting \cite{o1978curve}. 
Moreover, in kernel smoothing \cite{wand1994kernel}, the estimates are created from the weighted averages of nearby data points using the appropriate kernel, where the relevant weighting is a tapered window, i.e., it diminishes from the middle peak. While the traditional problem of moving mean tapering window design has been studied \cite{gokcesu2022smoothing}, we focus on its extension where the moving mean can be expressed as an auto-regressive model.


\section{Preliminaries}\label{sec:problem}

Let us have the observed samples 
\begin{align}
	\boldsymbol{y}=\{y_n\}_{n=1}^{N}.
\end{align}
We smooth $\boldsymbol{y}$ and create our smoothed signal
\begin{align}
	\boldsymbol{x}=\{x_n\}_{n=1}^{N}.
\end{align}

Traditionally, the most popular smoothing technique is the weighted moving average smoothing \cite{gokcesu2022smoothing}, where  $\boldsymbol{x}$ is created by passing $\boldsymbol{y}$ through a weighted moving average (weighted mean) filter $\boldsymbol{w}=\{w_k\}_{k=-M}^{M}$ of window size $2M+1\leq N$, where $\boldsymbol{w}$ is in a probability simplex, hence, the weights are positive, i.e.,
\begin{align}
	w_k\geq 0,&&k\in\{-M,\ldots,M\},
\end{align}
and they sum to $1$, i.e.,
\begin{align}
	\sum_{k=-M}^{M}w_k=1.
\end{align}
Let $N$ be odd, i.e., $N=2K+1$ for some natural number $K$. Then, without loss of generality, we can define the weight vector $\boldsymbol{w}$ for a window length of $N=2K+1$ with some trailing zeros, i.e.,
\begin{align}
	\tilde{w}_k=\begin{cases}
		w_k,&-M\leq k\leq M\\
		0,& \text{otherwise}
	\end{cases},
\end{align} 
for $k\in\{-K,\ldots,K\}$. Thus, without loss of generality, we can assume the window length is $N$.
The smoothed signal $\boldsymbol{x}$ is given by
\begin{align}
	x_n=\sum_{k=-K}^{K}w_ky_{n+k},\label{eq:wmean}
\end{align}  
where the mean filter is cyclic, i.e., 
\begin{align}
	y_{n'}=y_{n'+N}, &&\forall n'.\label{eq:ycyclic}
\end{align}
Note that the output $\boldsymbol{x}$ of this weighted moving average is a global minimizer of the following optimization problem (weighted cumulative cross error):
\begin{align}
	\min_{\boldsymbol{x}\in\Re^N}\sum_{n=1}^{N}\sum_{k=-K}^{K}w_k(y_{n+k}-x_n)^2,\label{eq:probx}
\end{align} 
since it is convex in $\boldsymbol{x}$ and the gradient is zero at its minimizer. Thus, given a weight vector $\boldsymbol{w}$ and the input $\boldsymbol{y}$, the solution of \eqref{eq:probx} is given by the $\boldsymbol{x}$ in \eqref{eq:wmean}. 

In this paper, instead of the objective in \eqref{eq:probx}, we study the following revised objective function to better enforce the smoothness in our output $\boldsymbol{x}$:

\begin{align}
F(\boldsymbol{x},\boldsymbol{y},\boldsymbol{w})=\sum_{n=1}^{N}\sum_{k=-K}^{K}w_k(z_{n,k}-x_n)^2,\label{eq:probxS}
\end{align} 
where 
\begin{align}
	z_{n,k}=\begin{cases}
		x_{n+k},& k\neq 0\\
		y_n, & k=0 
	\end{cases}.
\end{align}

We can extend the objective function to the following generalized form:

\begin{align}
	G(\boldsymbol{x,y,\alpha,\beta})=\sum_{n,k}\alpha_k(y_{n+k}-x_n)^2+\sum_{n,k}\beta_k(x_{n+k}-x_n)^2,\label{objG}
\end{align} 
for some nonnegative weights $\alpha_k$ and $\beta_k$, i.e.,
\begin{align}
	\alpha_k\geq& 0, &&\forall k,
	\\\beta_k\geq& 0, &&\forall k.
\end{align}

\begin{proposition}\label{thm:eqForm}
	The objective function in \eqref{objG} is equivalent to
	\begin{align*}
		G(\boldsymbol{x,y,\alpha,\beta})=H_0(\boldsymbol{y,\alpha})+H(\boldsymbol{x,\bar{y},\beta},A),
	\end{align*}
	where
	\begin{align*}
		H_0(\boldsymbol{y,\alpha})=&\sum_{n,k}\alpha_k(y_{n+k}-\bar{y}_n)^2,
		\\H(\boldsymbol{x,\bar{y},\beta},A)=&\sum_{n}A(\bar{y}_{n}-x_n)^2+\sum_{n,k}\beta_k(x_{n+k}-x_n)^2,
		\\\bar{y}_n=&\frac{\sum_{k}\alpha_ky_{n+k}}{A},
		\\A=&\sum_{k}\alpha_k.
	\end{align*}

	\begin{proof}
		The proof comes from the principle of orthogonality. We have
		\begin{align}
			\sum_{k}\alpha_k(y_{n+k}-x_n)^2=&\sum_{k}\alpha_k(y_{n+k}-\bar{y}_n+\bar{y}_n-x_n)^2,
			\\=&\sum_{k}\alpha_k\left[(y_{n+k}-\bar{y}_n)^2+(\bar{y}_n-x_n)^2\right]\nonumber
			\\&+\sum_{k}2\alpha_k(y_{n+k}-\bar{y}_n)(\bar{y}_n-x_n),
			\\=&\sum_{k}\alpha_k\left[(y_{n+k}-\bar{y}_n)^2+(\bar{y}_n-x_n)^2\right].
		\end{align}
		which concludes the proof.
	\end{proof}
\end{proposition}

\begin{remark}\label{thm:eqForm2}
	The objective function $H(\boldsymbol{x,\bar{y},\beta},A)$ in \autoref{thm:eqForm} is equivalent to the objective function $F(\boldsymbol{x,y,w})$ in \eqref{eq:probxS}, i.e.,
	\begin{align*}
		H(\boldsymbol{x,\bar{y},\beta},A)\leftrightarrow F(\boldsymbol{x,y,w})
	\end{align*}
	when
	\begin{align*}
		\bar{y}&\leftrightarrow y
		\\w_0&\leftrightarrow \frac{A}{A+B}
		\\w_k&\leftrightarrow\frac{\beta_k}{A+B}, &&k\neq0
	\end{align*}
\end{remark}

Thus, without loss of generality, we have the objective

\begin{align}
	\min_{\boldsymbol{x}\in\Re^N}\left[\sum_{n}w_0(\bar{y}_{n}-x_n)^2+\sum_{n}\sum_{k\neq 0}w_k(x_{n+k}-x_n)^2,\right],
\end{align} 
i.e., we want to minimize $F(\boldsymbol{x,\bar{y},w})$.

\section{Extracting the Smoothed Signal}\label{sec:fft}

In this section, we extract the smoothed signal $\boldsymbol{x}$.

\begin{lemma}\label{thm:convex}
	The objective function $F(\boldsymbol{x,\bar{y},w})$ in \eqref{eq:probxS} is convex in its argument $\boldsymbol{x}$.
	
	\begin{proof}
		Let $x_n=\lambda x_{n}^{(1)}+(1-\lambda)x_{n}^{(2)}$.
		We have
		\begin{align}
			\left(\bar{y}_{n}-x_n\right)^2=&\left(\bar{y}_{n}-\lambda x_{n}^{(1)}-(1-\lambda)x_{n}^{(2)}\right)^2,
			\\=&\left(\lambda\left(\bar{y}_{n}-x_{n}^{(1)}\right)+(1-\lambda)\left(\bar{y}_{n}-x_{n}^{(2)}\right)\right)^2,\nonumber
			\\\leq&\lambda\left(\bar{y}_{n}-x_{n}^{(1)}\right)^2+(1-\lambda)\left(\bar{y}_{n}-x_{n}^{(2)}\right)^2,
		\end{align}
		since the square function is convex. Similarly,
		\begin{align*}
			(x_{n+k}-x_n)^2=&\left[\lambda\left[ x_{n+k}^{(1)}-x_{n}^{(1)}\right]+(1-\lambda)\left[x_{n+k}^{(2)}-x_{n}^{(2)}\right]\right]^2,
			\\\leq&\lambda\left( x_{n+k}^{(1)}-x_{n}^{(1)}\right)^2+(1-\lambda)\left(x_{n+k}^{(2)}-x_{n}^{(2)}\right)^2.
		\end{align*}
		Thus, we have
		\begin{align}
			w_0&(\bar{y}_{n}-x_n)^2+\sum_{k\neq 0}w_k(x_{n+k}-x_n)^2
			\\\leq&\lambda\left[w_0\left(\bar{y}_{n}-x_n^{(1)}\right)^2+\sum_{k\neq 0}w_k\left(x_{n+k}^{(1)}-x_n^{(1)}\right)^2\right]\nonumber
			\\&+(1-\lambda)\left[w_0\left(\bar{y}_{n}-x_n^{(2)}\right)^2+\sum_{k\neq 0}w_k\left(x_{n+k}^{(2)}-x_n^{(2)}\right)^2\right].\nonumber
		\end{align}
		Hence, 
		\begin{align}
			F(\boldsymbol{x,\bar{y},w})\leq\lambda F(\boldsymbol{x^{(1)},\bar{y},w})+(1-\lambda)F(\boldsymbol{x^{(2)},\bar{y},w}),\nonumber
		\end{align}
		which concludes the proof.
	\end{proof}
\end{lemma}

We observe that because of the inherent symmetry in the objective function $F(\boldsymbol{x,\bar{y},w})$ due to the summation over $n$, we can manipulate $\boldsymbol{w}$ such that it is symmetric, i.e.,
\begin{align}
	w_k=w_{-k}, \forall k.
\end{align}

\begin{lemma}\label{thm:conv}
	The minimizer $\boldsymbol{x^*}$ of $F(\boldsymbol{x,\bar{y},w})$ is such that the circular convolution of $\boldsymbol{x}=\{x_n\}_{n=1}^N$ and $\boldsymbol{v}=\{v_k\}_{k=-K}^K$ gives $\boldsymbol{\bar{y}}=\{\bar{y}_n\}_{n=1}^N$, i.e., for positive $w_0$,
	\begin{align*}
		\boldsymbol{v}\circledast\boldsymbol{x}=\boldsymbol{\bar{y}},&&v_k=\begin{cases}
			\frac{2}{w_0}-1,& k=0\\
			-\frac{2w_k}{w_0},& k\neq0
		\end{cases}.
	\end{align*}
	\begin{proof}
		Since the objective function $F(\boldsymbol{x,\bar{y},w})$ is convex in $\boldsymbol{x}$ from \autoref{thm:convex}, its derivative at a minimizer $\boldsymbol{x_*}$ should be zero. Thus, since $w_k$ is symmetric and $\sum_{k}w_k=1$, we have
		\begin{align}
			(2-w_0)x_n=&w_0\bar{y}+\sum_{k\neq 0}w_kx_{n+k}+\sum_{k\neq 0}w_kx_{n-k},
			\\=&w_0\bar{y}+2\sum_{k\neq 0}w_kx_{n+k},
		\end{align}
		If $w_0=0$, any constant $\boldsymbol{x}$ is a solution. If $w_0>0$, we have
		\begin{align}
			\sum_{k}v_kx_{n+k}=\bar{y},
		\end{align}
		which concludes the proof.
	\end{proof}
\end{lemma}

To extract $\boldsymbol{x}$, we need to deconvolve $\boldsymbol{\bar{y}}$ with $\boldsymbol{v}$.

\begin{lemma}\label{thm:inv}
	The convolution by $\boldsymbol{v}$ is reversible, i.e., there exists a unique deconvolution.
	\begin{proof}
		Let $\mathcal{F}(\cdot)$ be the Fourier (DFT) operator. Because of the circular convolution, we have
		\begin{align}
			\mathcal{F}(\boldsymbol{v})\otimes\mathcal{F}(\boldsymbol{x})=\mathcal{F}(\boldsymbol{\bar{y}}),
		\end{align}
		where $\otimes$ denotes the Hadamard multiplication, i.e., element-wise multiplication.
		Let $\boldsymbol{V}$ be the DFT of $\boldsymbol{v}$, i.e., $\mathcal{F}(\boldsymbol{v})=\boldsymbol{V}$. Thus, the DFT of $\boldsymbol{v}$ is given by
		\begin{align}
			\boldsymbol{V}_n=&\sum_{k}v_ke^{-j\frac{2\pi kn}{N}},
			\\=&v_0+\sum_{k\neq 0}v_ke^{-j\frac{2\pi kn}{N}}
			\\=&v_0+\sum_{k=1}^Kv_k(e^{-j\frac{2\pi kn}{N}}+e^{j\frac{2\pi kn}{N}}),
		\end{align}
		because of the symmetry of $v_k$. Hence,
		\begin{align}
			\boldsymbol{V}_n=&v_0+\sum_{k=1}^K2v_k\cos\left(\frac{2\pi kn}{N}\right),
			\\=&1-\sum_{k=1}^{K}2v_k+\sum_{k=1}^K2v_k\cos\left(\frac{2\pi kn}{N}\right),
		\end{align}
		since $\sum_{k}v_k=1$. Thus,
		\begin{align}
			\boldsymbol{V}_n
			=&1-\sum_{k=1}^K2v_k\left(1-\cos\left(\frac{2\pi kn}{N}\right)\right),
			\\=&1+\sum_{k=1}^K 4\frac{w_k}{w_0}\left(1-\cos\left(\frac{2\pi kn}{N}\right)\right),
			\\\geq&1
		\end{align}
		since $w_k\geq 0$, $w_0>0$ and $\cos(\theta)\leq 1, \forall\theta$. Since all DFT points of $\boldsymbol{V}$ are nonzero, its element-wise multiplication is invertible, which concludes the proof.
	\end{proof}
\end{lemma}

\begin{theorem}
	Given the formulation in \eqref{objG} with the weights $\boldsymbol{\alpha}=\{\alpha_k\}_{k=-K}^K$ ($A=\sum_{k}\alpha_k$), $\boldsymbol{\beta}=\{\beta_k\}_{k=-K}^K$ ($B=\sum_{k}\beta_k$) and input $\boldsymbol{y}=\{y_n\}_{n=1}^K$; the minimizer $\boldsymbol{x_*}$ is
	\begin{align*}
		\boldsymbol{x_*}=\mathcal{F}^{-1}\left(\mathcal{F}\left(\boldsymbol{y}\right)\otimes\mathcal{F}(\boldsymbol{p})\oslash\mathcal{F}(\boldsymbol{v})\right),
	\end{align*}
where $\oslash$ denotes the Hadamard division, i.e., element-wise division, and
\begin{align*}
	p_k=\frac{1}{A}\alpha_k&& v_k=\begin{cases}
		\frac{A+2B}{A},& k=0\\
		-\frac{2\beta_k}{A},& k\neq0
	\end{cases}.
\end{align*}
	\begin{proof}
		From \autoref{thm:eqForm}, we have
		\begin{align}
			\mathcal{F}(\boldsymbol{\bar{y}})=\mathcal{F}(\boldsymbol{y})\otimes\mathcal{F}\left(\boldsymbol{p}\right).
		\end{align}
		 From \autoref{thm:conv} and \autoref{thm:inv}, we have		
		\begin{align}
			\mathcal{F}(\boldsymbol{x_*})=\mathcal{F}(\boldsymbol{\bar{y}})\oslash\mathcal{F}(\boldsymbol{v}),
		\end{align} 
		which concludes the proof.
	\end{proof}
\end{theorem}

\section{Weight Design}

Given the general objective in \eqref{objG}, after extracting the smoothed signal $\boldsymbol{x}$ as in \autoref{sec:fft}, the question remains as to which weights are best suited to be chosen for optimal smoothing. Let 
\begin{align}
	J(\boldsymbol{y,\theta})=\min_{\boldsymbol{x}}G(\boldsymbol{x,y,\alpha,\beta}),
\end{align}
where $\boldsymbol{\theta}=(\boldsymbol{\alpha,\beta})$ cumulatively represents the weights. 

To model the optimal smoothing we can minimize the objective function $J(\boldsymbol{y,\theta})$. Since the weights $\boldsymbol{\alpha,\beta}$ are nonnegative, i.e., $\boldsymbol{\alpha}\geq 0$ and $\boldsymbol{\beta}\geq 0$, the minimizer weights $\boldsymbol{\theta_*}$ would be all-zero because the objective is always nonnegative. To circumvent this, we need to restrict the weights.
\begin{remark}
	We use the most popular regularization, which is to limit the sum of weights to $1$, i.e., $\abs{\boldsymbol{\theta}}_1=1$ and represents a probability distribution. Hence, $A+B=1$.  
\end{remark}

\begin{lemma}
	$J(\boldsymbol{y,\theta})$ is concave in $\boldsymbol{\theta}$.
	\begin{proof}
		$G(\boldsymbol{x,y,\alpha,\beta})$ is linear in the weights $\boldsymbol{\theta}=(\boldsymbol{\alpha,\beta})$, i.e., 
		\begin{align}
			G(\boldsymbol{x,y},\sigma\boldsymbol{\alpha},\sigma\boldsymbol{\beta})=\sigma G(\boldsymbol{x,y},\boldsymbol{\alpha},\boldsymbol{\beta}),
		\end{align}
		for $\forall \sigma\in\Re$, and $J(\boldsymbol{y,\theta})$ is its minimization with respect to $\boldsymbol{x}$; which concludes the proof from \cite{gokcesu2022smoothing}.
	\end{proof}
\end{lemma}

Let $\tilde{\alpha}_n=\alpha_n/A$ and $\tilde{\beta}_n=\beta_n/B$. We can see that each of $\boldsymbol{\theta_\alpha}=(\boldsymbol{\tilde{\alpha},0})$ and $\boldsymbol{\theta_\beta}=(\boldsymbol{0,\tilde{\beta}})$ are probability distributions themselves, where the original weighting distribution is given by their convex combination $\boldsymbol{\theta}=A\boldsymbol{\theta_\alpha}+B\boldsymbol{\theta_\beta}$.
An optimal set of weights for a concave function in a convex polytope is one of its vertices \cite{gokcesu2022smoothing}. Hence, the unrestrained minimization results in either all zeros for $\boldsymbol{\alpha}$ or all zeros for $\boldsymbol{\beta}$. If $\boldsymbol{\beta}$ is all-zeros, the smoothed signal will be equal to the weighted mean of $\boldsymbol{y}$ with $\boldsymbol{\alpha}$. If $\boldsymbol{\alpha}$ is all-zeros, the smoothed signal will be an all-constant signal, which will result in the minimum error (trivially zero). Hence, increasing $B$ directly decreases the error.

To circumvent these trivialities, we need to restrain $A$ and $B$. Given fixed $A$ and $B$ (where $A+B=1$), we have
\begin{align}
	p_k=&\frac{\alpha_k}{A}, \\q_k=&\frac{\beta_k}{B},
	\\v_k=&\begin{cases}
		\frac{2}{A}-1,& k=0\\
		-\frac{2B}{A}q_k,& k\neq0
	\end{cases}
\end{align}
Unrestrained optimization of $\boldsymbol{p}$ and $\boldsymbol{q}$ will again result in a trivial solution. To this end, we restrict them to symmetric tapering window functions such that
\begin{align}
	p_{-1}=&p_0=p_1
	\\p_k=&p_{-k}
	\\p_{k_1}\geq& p_{k_2}, &&\abs{k_1}<\abs{k_2}
\end{align}
Similarly, for $q_k$,
\begin{align}
	q_k=&q_{-k}
	\\q_{k_1}\geq& q_{k_2}, &&\abs{k_1}<\abs{k_2}
\end{align}

\begin{remark}
	$A=1/3$ and $B=2/3$ is a natural choice to guarantee tapering behavior for $w_k$ as in \autoref{thm:eqForm2} while making $A$ as small as possible. 
\end{remark}
We observe that both $p_k$ and $q_k$ can be written in terms of a convex mixture of bandlimited uniform distributions because of the tapering. The convex mixture of uniform distributions create a convex polytope. Because of the concavity of the objective, one of the vertices is the solution \cite{gokcesu2022smoothing}.
\begin{remark}
	Given a window length limitation $O(L)$ for both $p_k$ and $q_k$; there exist $O(L^2)$ vertices in this polytope. We can try all of these vertices to find the optimizer, which will take $O(L^2N\log N)$ time. If $L$ is polynomially dependent on $\log N$, then the time complexity will be quasilinear, i.e., $\tilde{O}(N)$.
\end{remark}
\begin{remark}
	For increased efficiency, we may also limit the distributions to equal to each other, which would result in a polytope with $O(L)$ vertices. Consequently, the time complexity will be $O(LN\log N)$ and similarly $\tilde{O}(N)$ for poly-logarithmic $L$. 
\end{remark}
\begin{remark}
	If we do not want to limit the choice of $A$ and $B$, we may do a heuristic optimization in a cascade manner. We can first optimize $\boldsymbol{p}$ using $H_0(\boldsymbol{y,\alpha})$ in \autoref{thm:eqForm}, which will give us $\boldsymbol{\bar{y}}$.
	From $\boldsymbol{\bar{y}}$, we can optimize the weighting $\boldsymbol{q}$ using $H(\boldsymbol{x,\bar{y},\beta},A)$. However, only the tapering limitation will trivially result in a wide-bandwidth weighting. To this end, we can smooth with the narrowest bandwidth AR filter (length $3$), i.e., $w_{-1}=w_0=w_1=1/3$. The whole process to find $\boldsymbol{x}$ will take $O(N\log N)$ linearithmic time \cite{gokcesu2022smoothing}.
\end{remark}

\begin{remark}
	In the cascade, the first filter $\boldsymbol{p}$ is a weighted mean filter. While the second AR filter, the deconvolution by $\boldsymbol{v}$, has a curious window shape. While it is also a unimodal window function, it has exponential tapering at its asymptotes. Let us define the deconvolution with $\boldsymbol{u}$, whose convolution with $\boldsymbol{v}$ will be the dirac-delta. Hence, the elements of $\boldsymbol{u}$ will have a recurrence relation (linear difference equation) with the coefficients $\boldsymbol{v}$. Consider the characteristic polynomial implied by $\boldsymbol{v}$, which is
	\begin{align}
		\Lambda(r)=\sum_{k=-K}^{K}v_kr^k.
	\end{align}
	Since $v_k$ is symmetric, we have
	\begin{align}
		\Lambda(r)= v_0+\sum_{k=1}^{K}v_k(r^k+r^{-k}).
	\end{align}
	From \autoref{thm:inv}, we know that this polynomial does not have complex roots on the unit circle. Moreover, we have
	\begin{align}
		\lim_{r\rightarrow0^{+}}\Lambda(r)=-\infty,&& 	\Lambda(1)=v_0+\sum_{k=1}^{K}2v_k=1>0,
	\end{align}
	since $v_k\leq 0$ for $k\neq 0$. Furthermore, since
	\begin{align}
		\Lambda'(r)=\sum_{k=1}^{K}kv_kr^{-1}(r^{k}-r^{-k})>0, &&r\in(0,1),
	\end{align} 
	we have a unique root in $(0,1)$. Thus, the deconvolution window will have near exponentially decreasing tails. 
\end{remark}


\bibliographystyle{IEEEtran}
\bibliography{double_bib}

\end{document}